\documentclass[
]{ceurart}

\usepackage{listings}
\begin{document}

\copyrightyear{2025}
\copyrightclause{Copyright for this paper by its authors.
  Use permitted under Creative Commons License Attribution 4.0
  International (CC BY-NC-ND 4.0).}

\conference{HOOC Workshop - Higher Order Opportunities and Challenges, 11th August 2025 at RWTH Aachen, Germany}

\title{A Remedy for Over-Squashing in Graph Learning via Forman‐Ricci Curvature based Graph-to-Hypergraph Structural Lifting}


\author{Michael Banf }[email=michael.banf@perelyn.com,]
\cormark[1]
\author{Dominik Filipiak}
\author{Max Schattauer}
\author{Liliya Imasheva}

 \address{Perelyn GmbH, Reichenbachstraße 31, 80469 München, Germany }

\cortext[1]{Corresponding author.}

\begin{abstract}
Graph Neural Networks are highly eﬀective at learning from relational data, leveraging node and edge features while maintaining the symmetries inherent to graph structures. However, many real-world systems—such as social or biological networks—exhibit complex interactions that are more naturally represented by higher-order topological domains. The emerging field of Geometric and Topological Deep Learning addresses this challenge by introducing methods that utilize and benefit from higher-order structures. Central to TDL is the concept of lifting, which transforms data representations from basic graph forms to more expressive topologies before the application of GNN models for learning. In this work, we propose a structural lifting strategy using Forman-Ricci curvature, which defines an edge-based network characteristic based on Riemannian geometry. Curvature reveals local and global properties of a graph such as to identify a network’s backbones, i.e., coarse, structure-preserving graph geometries that form connections between major communities — most suitably represented as hyperedges to model information flows between clusters across large distances in the network. To this end, our approach provides a remedy to the problem of information distortion in message passing across long distances and graph bottlenecks—a phenomenon known in graph learning as over-squashing. 
\end{abstract}


\maketitle

\section*{The challenge: over-squashing in graph learning}\label{section:challenge}
\begin{figure}
\includegraphics[width=\textwidth]{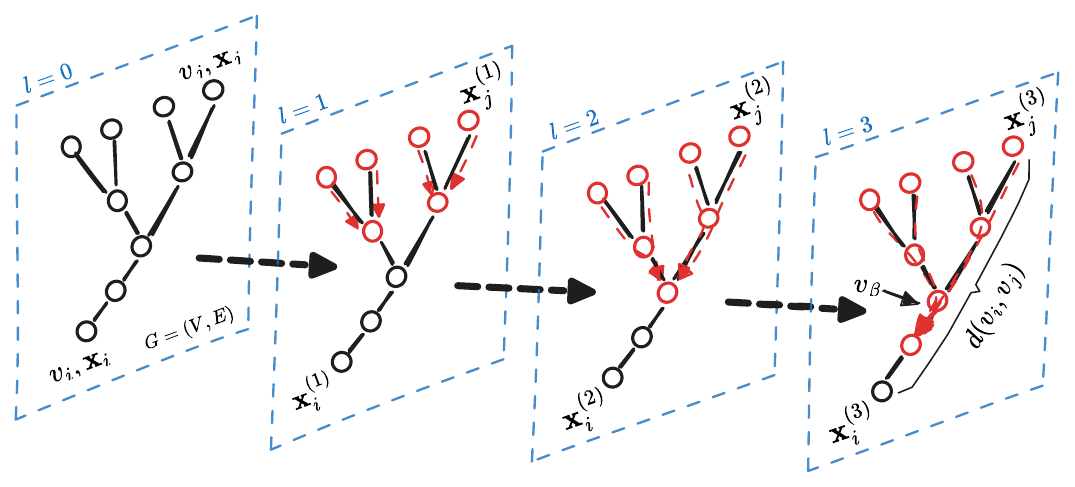}
\caption{Layer-wise information propagation along edges of $G$. Due to $G$'s structure, node $v_\beta$ forms information bottleneck. In addition, as $d(v_i,v_j) = 4$ but number of layers $l = 3$, information from $v_j$ never reaches $v_i$.} \label{fig:bottleneck}
\end{figure}

\noindent Graph Neural Network (GNN) architectures often are built on the message-passing paradigm, where nodes exchange information along a graph's edges. In such a framework, the input graph plays a dual role: it encodes structural relationships and serves as the backbone for feature propagation. However, recent research ~\cite{Topping2022, Attali2024, Akansha2025} shows that certain graph structures are inherently ill-suited for effective message passing. Specifically, Message Passing Neural Networks (MPNNs), such as Graph Convolutional Neural Networks (GCN) or Graph Attention Networks (GAT) ~\cite{Nguyen2024}, struggle in capturing long-range dependencies, where model outputs rely on interactions between distant nodes ~\cite{Dwivedi2022}. In such scenarios, messages from distant nodes must traverse a limited number of paths and be compressed into fixed-size representations. This results in bottlenecks in the information flow (see illustration in figure~\ref{fig:bottleneck}), where critical signals may be lost, a phenomenon known as information over-squashing ~\cite{Dwivedi2022, Attali2024, Akansha2025}. To this end, over-squashing can be quantified as the lack of sensitivity of an MPNN at a given node to the input features of another remote node \cite{Topping2022, Black2023}. Consider a graph $G = (V, E)$ with node features $X \in \mathbb{R}^{n \times m}$, given $n$ nodes and feature size $m$. $\mathbf{x_{i}} \in \mathbb{R}^m$ denotes a row in $X$ corresponding to a node $v_i \in V$. A general MPNN updates node features by iteratively, i.e. layer-wise, aggregating features of nodes in the neighborhood of $v_i$ according to:

\begin{equation}
\mathbf{x}_{i}^{(0)}:=\mathbf{x}_{v_i}, \quad \mathbf{x}_i^{(l)} = \phi^{(l)} \left( \mathbf{x}_i^{(l - 1)}, \bigoplus_{j \in \mathcal{N}(v_i)} \, \psi^{(l)}\left(\mathbf{x}_i^{(l - 1)}, \mathbf{x}_j^{(l - 1)},\mathbf{e}_{i, j}\right) \right)
\end{equation}\label{eq:gnn}

\newpage

\noindent Here, $\mathbf{x}^{(l)}_i$ and $\mathbf{x}^{(l-1)}_i$ denote the $l$-th and $l$-1-th layers' node features representations of $v_i$, $\bigoplus$ is a differentiable, permutation invariant function, e.g., sum or mean, $\psi$ and $\phi$ are differentiable message aggregation and node update functions, $\mathcal{N}(v_i)$ is the node neighborhood of $v_i$ and $\mathbf{e}_{i, j}$ denote some (optional) edge features from node $v_j$ to $v_i$. The receptive field $\mathcal{R}^{l}$ at the $l$-th layer is defined as $\mathcal{R}^{l} \{v_i, v_j \in V: d(v_i,v_j) \leq l\}$, where $d(v_i,v_j)$ represents the shortest path distance in $G$ between $v_i$ and $v_j$. The Jacobian $\partial \mathbf{x}_i^{(l)} / \partial \mathbf{x}_j$ quantifies the sensitivity of representation $\mathbf{x}_i^{(l)}$ at $v_i$ to a specific input feature $\mathbf{x}_j$ at $v_j$ within $\mathcal{R}_{l}$. Small absolute values of this Jacobian are indicative of poor information propagation, or over-squashing, i.e. the incapacity of $\mathbf{x}_i^{(l)}$ to be influenced by $\mathbf{x}_j$ at a distance of $l$. For illustration, let's take the graph $G$ in figure~\ref{fig:bottleneck}. Due to the specific structure of $G$, i.e. the exponential grows of $R_l$ as $d$ increases, in particular node $v_\beta$ forms an information bottleneck. In addition, since the maximum number of layers of the MPNN is $l = 3$, information from $v_j$ never reaches $v_i$, since $d(v_i,v_j) = 4$ and, hence, $d(v_i,v_j) \not \leq l$. In other words $v_j$ is outside the receptive field $\mathcal{R}_{l}$ of $v_i$. Notably,  $\partial \mathbf{x}_i^{(l)} / \partial \mathbf{x}_j$ can be upper-bounded using powers of the $G$'s normalized adjacency matrix, revealing that the structure of $G$ itself governs the extent of over-squashing \cite{Topping2022, Black2023}. \\

\section*{The metric: curvature on graphs}\label{section:metric}
\begin{figure}
\includegraphics[width=\textwidth]{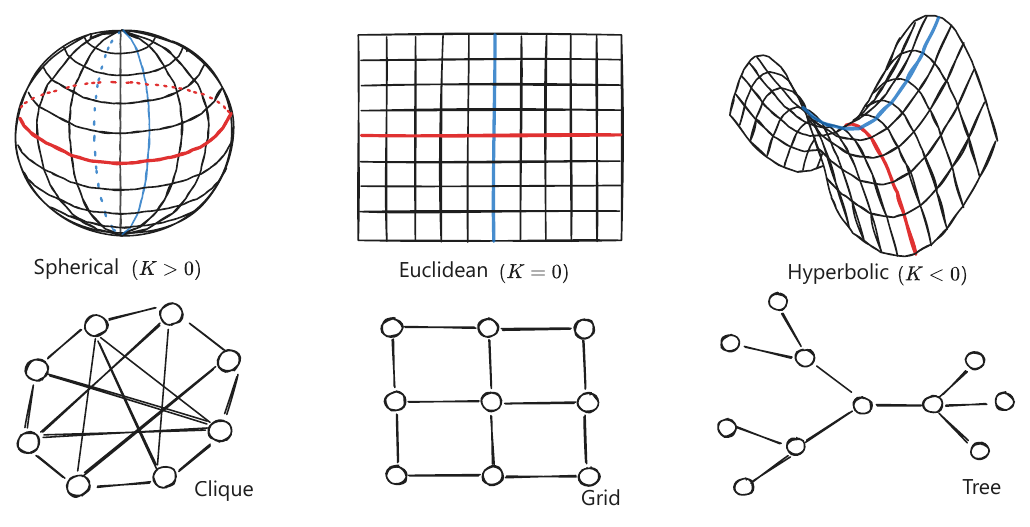}
\caption{In differential geometry, curvature on manifolds, such as Gaussian curvature $K$, relates to information propagation. Replacing geodesics for edges, one can find discrete graph analogous for different types of curvature.}\label{fig:curvature} 
\end{figure}

\noindent To gain a deeper understanding of the over-squashing phenomenon and its relationship to the graph structure itself, analyzing the fine-grained geometric properties of the graph has been proposed \cite{Topping2022}. One such property would be curvature, which - in its continuous form in differential geometry - measures how space bends locally. In two dimensions, Gaussian curvature $K$ defines whether a surface is flat (Euclidean), positively curved (e.g. a sphere), or negatively curved (e.g. a saddle). Being the product of a surface's principal curvatures, $K$ denotes an intrinsic property of a surface independent on a specific embedding in any higher dimensional space (figure~\ref{fig:curvature}). In higher dimensions, concepts such as Ricci curvature or sectional curvature generalize this idea. \\

\noindent Curvature thus serves as a bridge between geometry and topology, enabling local information to yield global insight. Recently such notions have been extended to the discrete domain of graphs \cite{Topping2022}. As an analogy of geodesic dispersion - i.e. whether two parallel geodesics starting from nearby points converge ($K > 0$), remain parallel ($K = 0$), or diverge ($K < 0$) - on graphs, consider an edge with two edges starting at its respective endpoints. In a discrete spherical geometry (e.g. a clique graph), the edges would meet at another node to form a triangle. In a discrete Euclidean geometry (e.g. a grid), the edges would stay parallel and form a rectangle or grid, whereas in a discrete hyperbolic geometry (e.g. a tree), the mutual distance of the edge endpoints would increase. \\

\noindent To measure curvature on graphs, a variety of metrics have been proposed, predominantly Ollivier-Ricci \cite{Sia2019} as well as several variations of Forman-Ricci curvature \cite{Samal2018, Topping2022, Saucan2018, Attali2024}. In particular, Ollivier-Ricci curvature \cite{Ollivier2009} translates previously described notions of positive and negative curvature directly to the discrete domain \cite{Topping2022}. It has also been shown that both Ollivier-Ricci and Forman-Ricci tend to correlate to a certain extent w.r.t the local and global properties they evaluate \cite{Samal2018}. This has practical implications w.r.t to using Forman-Ricci curvature instead of Ollivier-Ricci curvature for more eﬃcient computation scalable to large network structures. However, it has to be noted that the classical, i.e. non-augmented, purely edge-based form of Forman-Ricci curvature on graphs \cite{Saucan2018, Samal2018, Weber2017}, to a certain degree, biases towards negative values, hence lacks the intuition of Ollivier-Ricci curvature. Concretely, for Forman-Ricci curvature highly negative values tend to describe edges within clusters whereas values equal or slightly smaller than zero indicate bridges between clusters. 

\section*{The remedy: curvature guided graph-to-hypergraph structural lifting}
\noindent As a consequence of identifying graph bottlenecks via curvature, local graph rewiring has been proposed \cite{Attali2024}, i.e. adding edges around bottleneck structures, e.g. curvature-guided via discrete Ricci Flows, inspired by how continuous Ricci flow evolves the metric of a Riemannian manifold proportionally to the Ricci curvature \cite{Topping2022}. However, one fundamental restriction of graphs is their limitation to pairwise relations. To this end, methodologies from the field of Topological Deep Learning overcome such constraints by encoding higher-order relationships through combinatorial and algebraic topology concepts, which enable more complex relational representations via part-whole and set-types relations \cite{Bernárdez2024}. \\

\begin{figure}
\includegraphics[width=\textwidth]{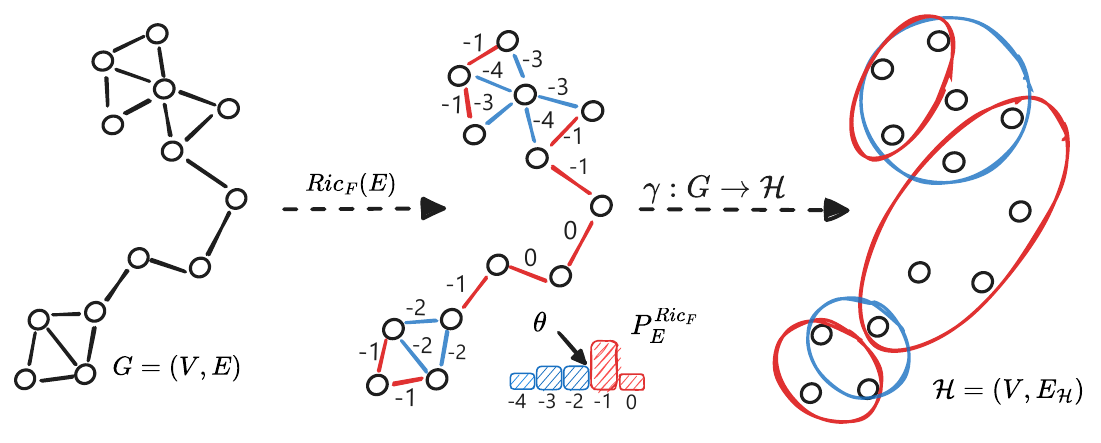}
\caption{Forman-Ricci curvature based structural lifting $\gamma: G \rightarrow \mathcal{H}$ given $\theta = -1$ for an unweighted graph $G$. Red and blue hyper-edges correspond to ${Ric_F}(e) \geq \theta$ and ${Ric_F}(e) < \theta$, respectively. }\label{fig:forman} 
\end{figure}

\noindent Here we propose a remedy to the over-squashing problem that first exploits curvature, i.e. Forman-Ricci curvature, as a metric to, in particular, identify a network's structure-preserving, coarse geometry - i.e. its backbones, which lend themselves specifically to model information flows across wide areas of the network - and subsequently performs a structural lifting \cite{Hajij2023, Papillon2023} as a mapping based on connections within the original input graph $G$ to form a hypergraph $\mathcal{H}$, thereby treating identified backbones as hyperedges, virtually shortening the distance between potentially relevant nodes. We evaluate the Forman-Ricci curvature ${Ric_F}(e)$ of each edge $e$ in $G$, which, for an undirected edge $e \in E$ between nodes $v_i, v_j \in V$ with weights $w(v_i), w(v_j)$, is defined as \cite{Samal2018}:  \\

\begin{equation}\label{eq:ric}
    {\rm Ric_F}(e) = \omega (e) \left( \frac{\omega (v_i)}{\omega (e)} +  \frac{\omega (v_j)}{\omega (e)}  - \sum_{\substack{e_{v_i}\ \sim\ e \\ \ e_{v_j}\ \sim\ e}} \left[\frac{\omega (v_i)}{\sqrt{\omega (e) \omega (e_{v_i})}} + \frac{\omega (v_j)}{\sqrt{\omega (e) \omega (e_{v_j})}} \right] \right)
\end{equation}

\noindent where $e_{v_i}, e_{v_j}$ are the edges connected to nodes $v_i$ and $v_j$. For an unweighted network graph, i.e. all edge weights $w = 1$, equation \ref{eq:ric} reduces to ${Ric_F}(e) = 4 - \deg(v_i) - \deg(v_j)$. Thus ${Ric_F}(e) << 0$ if both $v_i$ and $v_j$ have high degrees. In contrast, ${Ric_F}(e) = 0$ if $v\deg(v_i) = \deg(v_j) = 2$. The intuition behind this definition, is to capture how fast edges spread in different directions. In particular, edges with high negative curvature, i.e. ${Ric_F}(e) << 0$, should play a special role for the spreading out and hence for, e.g., information dispersal in a network, whereas edges with ${Ric_F}(e) \approx 0$ rather constitute long-range interactions, bottlenecks, interfaces or boundaries (see figure~\ref{fig:forman}). \\

\noindent Subsequently, we use ${Ric_F}(e)$ values per each $e$ to define a structural lifting $\gamma: G \rightarrow \mathcal{H}$, to map $G = (V, E)$, i.e. in particular all nodes $V$ in $G$ to $\mathcal{H} = (V, E_\mathcal{H})$. $\mathcal{H}$ generalizes $G$ by allowing hyperedges to connect any number of nodes \cite{Hajij2023, Papillon2023}, thereby exhibiting set-type relationships that lack an explicit notion of hierarchy. More formally, given $G = (V, E)$, a hypergraph $\mathcal{H}$ denotes the pair $(V, \mathcal{E}_{\mathcal{H}})$, where $\mathcal{E}_{\mathcal{H}}$ is a non-empty subset of the powerset $\mathcal{P}(V) \setminus \{\emptyset\}$, with elements of $\mathcal{E}_{\mathcal{H}}$ being hyperedges. As auxiliary variable we define a custom hyperparameter $\theta$ that serves to separate two types of hyperedges, i.e. one denoting long-range interactions, bottlenecks, interfaces or boundaries, and the other denoting clusters. Note that another way to define $\theta$ is based on $P^{Ric_F}_E$, i.e. the (empirical) distribution of ${Ric_F}(e)$ of all $e \in E$ in $G$ (as illustrated in figure~\ref{fig:forman}). Based on the quantile function $Q_{P^{Ric_F}_E}: [0,1] \rightarrow \mathbb{R}$ we can define $Q_{P^{Ric_F}_E}(p) = \theta$, with $p$ denoting some chosen probability. For example, if $p = 0.9$ then $\theta$ would correspond to the curvature value for the $0.9$-th quantile of $P^{Ric_F}_E$. \\

\noindent We then define a hyperedge $e_{S, \theta} \in \mathcal{E}_{\mathcal{H}}$ in $\mathcal{H}$ where $S$ denotes a subset of $V$ in $G$. For each node $v_i \in S$, $v_i$ is an endpoint to an edge $e$ in $G$ with ${Ric_F}(e) < \theta$ (or ${Ric_F}(e) \geq \theta$ respectively), and $\exists v_j \in S$, with $v_j \in \mathcal{N}_1(v_i)$, i.e. $v_i, v_j$ are connected via $e$ in $G$. Figure~\ref{fig:forman} illustrates $\gamma: G \rightarrow \mathcal{H}$ given $\theta = -1$.

\section*{Experiments}
\noindent We evaluate our approach on graph classification tasks based on multiple, well-known graph learning benchmark datasets, as well as a large molecular property prediction dataset (see table 1). Per model architecture and experimental dataset, results are averaged across four metrics, i.e. precision, recall, auroc and accuracy. Using two common graph neural network models (GCN, GAT), we find the lifted datasets to consistently, i.e., in 75 \% (6 out of 8) of test cases, improve model performance compared to applying the same models to the non-lifted datasets directly. Intriguingly, even multiple higher-order interaction-encoding hypergraph network models (EDGNN, AllSetTransformer, UniGNN2) demonstrate increased performance, i.e., in 80 \% (10 out of 12) of test cases, when complemented with a preceding lifting of the datasets. 
\\

\begin{table}[h!]
  \centering
    \label{tab:sample}
        \caption{Performances on original and lifted datasets across several graph and hypergraph neural network architectures.}
  
    \begin{tabular}{llllll}
        \toprule
        \multirow{2.5}{*}{Dataset / Method} \\ 
            & GCN & GAT & AST & EDGNN & UniGNN2 \\
            NCI1 & 0.71 $\pm$ 0.03  & 0.73 $\pm$ 0.04 & 0.65 $\pm$ 0.07 & 0.70 $\pm$ 0.06  & 0.63 $\pm$ 0.08 \\
            NCI1 ($Ric_{F}$) & \textbf{0.72} $\pm$ \textbf{0.03}  & 0.73 $\pm$ 0.04 & \textbf{0.68} $\pm$ \textbf{0.05} & \textbf{0.71} $\pm$ \textbf{0.04} & \textbf{0.72} $\pm$ \textbf{0.03} \\
            \midrule
            NCI109 & 0.70 $\pm$ 0.03 & 0.71 $\pm$ 0.04 & 0.70 $\pm$ 0.06 & 0.68 $\pm$ 0.09  & 0.67 $\pm$ 0.09 \\
            NCI109 ($Ric_{F}$) & \textbf{0.71} $\pm$ \textbf{0.03} & \textbf{0.72} $\pm$ \textbf{0.04} & 0.69 $\pm$ 0.07 & \textbf{0.69} $\pm$ \textbf{0.07} & \textbf{0.70} $\pm$ \textbf{0.05} \\
            \midrule
        OGBG Molhiv & 0.78 $\pm$ 0.14 & 0.77 $\pm$ 0.17 & 0.78 $\pm$ 0.20 & 0.76 $\pm$ 0.17 &  0.78 $\pm$ 0.15 \\
            OGBG Molhiv ($Ric_{F}$) & \textbf{0.79} $\pm$ \textbf{0.13} & 0.77 $\pm$ 0.17 & \textbf{0.79} $\pm$ \textbf{0.19} & \textbf{0.78} $\pm$ \textbf{0.16} & 0.78 $\pm$ 0.16 \\
            \midrule
            PROTEINS & 0.69 $\pm$ 0.04 & 0.69 $\pm$ 0.04 & 0.66 $\pm$ 0.08 & 0.67 $\pm$ 0.07 & 0.69 $\pm$ 0.06\\
            PROTEINS ($Ric_{F}$) & \textbf{0.70} $\pm$ \textbf{0.02} & \textbf{0.70} $\pm$ \textbf{0.04} & \textbf{0.70} $\pm$ \textbf{0.07} & \textbf{0.73} $\pm$ \textbf{0.04} & \textbf{0.70} $\pm$ \textbf{0.09} \\
        \bottomrule
    \end{tabular}
  
\end{table}

\section*{Implementation}
An earlier version of our curvature-based structural lifting has been made available as part of TopoBench \cite{Telyatnikov2024} a modular Python library designed to standardize benchmarking and accelerate research in Topological Deep Learning.


%

\end{document}